\documentclass[conference]{IEEEtran}
\IEEEoverridecommandlockouts
\pdfoutput=1
\usepackage{cite}
\usepackage{amsmath,amssymb,amsfonts}
\usepackage{algorithm}
\usepackage{algorithmic}
\usepackage{bbm}
\usepackage{booktabs}
\usepackage{pifont}
\usepackage{graphicx}
\usepackage{textcomp}
\usepackage{multirow}
\usepackage[table]{xcolor}
\definecolor{LightBlue}{rgb}{0.9,0.94,1}
\def\BibTeX{{\rm B\kern-.05em{\sc i\kern-.025em b}\kern-.08em
    T\kern-.1667em\lower.7ex\hbox{E}\kern-.125emX}}

\makeatletter
\renewcommand{\maketag@@@}[1]{\hbox{\m@th\normalsize\normalfont#1}}%
\makeatother

\begin{document}

\title{\textbf{DH-VTON: Deep Text-Driven Virtual Try-On via Hybrid Attention Learning}}

\author{\IEEEauthorblockN{Jiabao Wei}
\IEEEauthorblockA{\textit{School of Computer Science and Technology} \\
\textit{Beijing Institute of Technology}\\
Beijing, China \\
weijiabao@bit.edu.cn}
\and
\IEEEauthorblockN{Zhiyuan Ma}
\IEEEauthorblockA{\textit{Department of Eletronic Engineering} \\
\textit{Tsinghua University}\\
Beijing, China \\
mzyth@tsinghua.edu.cn}
}

\maketitle

\begin{abstract}
Virtual Try-ON (VTON) aims to synthesis specific person images dressed in given garments, which recently receives numerous attention in online shopping scenarios. 
Currently, the core challenges of the VTON task mainly lie in the fine-grained semantic extraction (\emph{i.e., deep semantics}) of the given reference garments during depth estimation and effective texture preservation when the garments are synthesized and warped onto human body. To cope with these issues, we propose DH-VTON, a deep text-driven virtual try-on model featuring a special hybrid attention learning strategy and deep garment semantic preservation module. By standing on the shoulder of a well-built pre-trained paint-by-example (\emph{abbr. PBE}) approach, we present our DH-VTON pipeline in this work. Specifically, to extract the deep semantics of the garments, we first introduce InternViT-6B as fine-grained feature learner, which can be trained to align with the large-scale intrinsic knowledge with deep text semantics (\emph{e.g., ``neckline'' or ``girdle''}) to make up for the deficiency of the commonly adopted CLIP encoder. Based on this, to enhance the customized dressing abilities, we further introduce \underline{G}arment-\underline{F}eature \underline{C}ontrolNet \underline{Plus} (\emph{abbr. GFC+}) module and propose to leverage a fresh hybrid attention strategy for training, which can adaptively integrate fine-grained characteristics of the garments into the different layers of the VTON model, so as to achieve multi-scale features preservation effects. 
Extensive experiments on several representative datasets demonstrate that our method outperforms previous diffusion-based and GAN-based approaches, showing competitive performance in preserving garment details and generating authentic human images. 
\end{abstract}


\section{Introduction}
Image-based virtual try-on (VTON) has recently attracted significant interests in generative research community~\cite{ma2024neural} with the increasing popularity of online shopping\cite{tpd,MGD,stableviton,ootdiffusion,IDMVTON,catDM,viton-hd,HRviton,DCI-VTON}. 
Despite significant progress having been witnessed, the existing VTON models still face several critical issues. One key issue lies in that the given garments must be naturally deformed to fit the target person's pose and body shape. Based on this, the other key issue is the patterns and texture details of the deformed garments need to be fine-grained preserved. 

To address the above two critical issues, existing image-based VTON approaches generally can be categorized into two categories: warping-based and warping-free approaches. \textbf{a)} The former~\cite{viton,viton-hd,cvton,HRviton,clothflow,DCI-VTON,zflow,size,LaDI-VTON,GPvton,issenhuth2020not,yang2022full,yu2019vtnfp} typically perform garment warping before image synthesis via GANs\cite{GAN} or LDMs\cite{LDM}. 
Early approaches primarily rely on GANs\cite{GAN} as image generator and tentatively reduce the mismatch between the warped garment and the target person such as in VITON-HD\cite{viton-hd}, HR-VITON\cite{HRviton} and GP-VTON\cite{GPvton}. Afterwards, researchers have considered leveraging LDMs\cite{LDM} instead of GANs\cite{GAN} as image generator due to their impressive generation capabilities\cite{DCI-VTON,LaDI-VTON}. 
Specifically, DCI-VTON\cite{DCI-VTON} and LaDI-VTON~\cite{LaDI-VTON} are two representative works by utilizing LDMs\cite{LDM} to merge the warped garment onto the target person. 
However, the main disadvantage of the warping-based approaches is the artifacts produced by the garment warping process, which may be difficult to eliminate during image synthesis. 
Furthermore, existing garment deformation methods such as TPS\cite{TPS}, STN\cite{STN}, and FlowNet\cite{flownet} basically all lack well customized dressing abilities under giving the various postures as conditions. 

\begin{figure}[!t]
\vspace{-0.4cm}
\centerline{\includegraphics[scale=0.88]{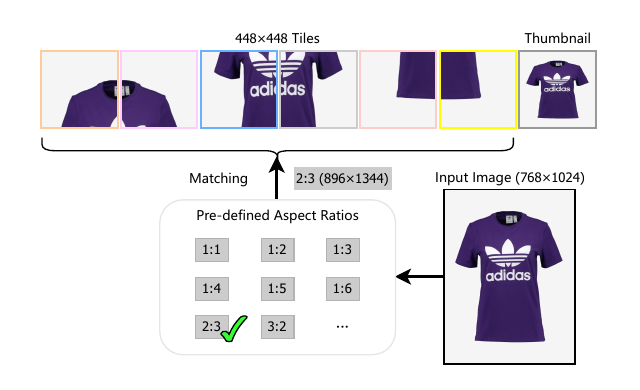}}
\setlength{\abovecaptionskip}{-0.2cm}
\caption{\textbf{Demonstration of dynamic high-resolution capabilities.} InternViT-6B-448px-V1-5\cite{internvit-6b} dynamically match an optimal aspect ratio from pre-defined ratios, dividing the image into tiles of $448\times448$ pixels and creating a thumbnail for global context.}
\label{InternViT}
\vspace{-0.4cm}
\end{figure}

\begin{figure*}[!h]
  \centering
  \includegraphics[width=\textwidth]{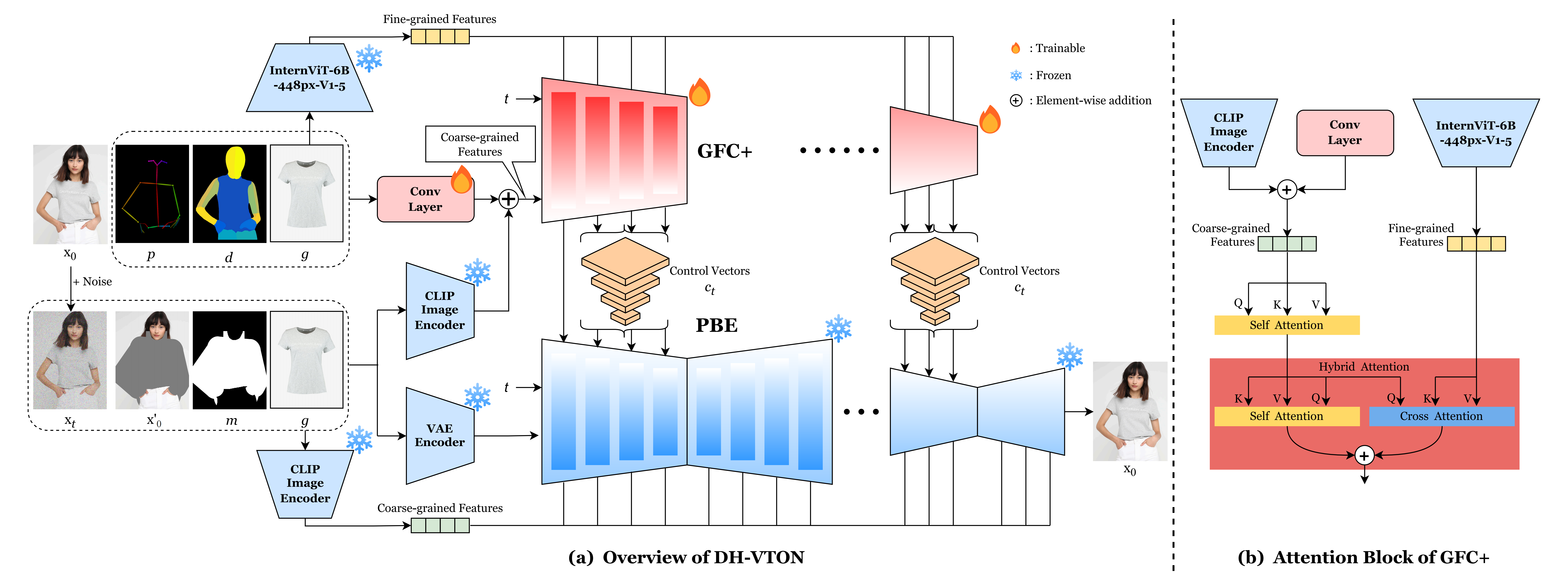}
  \setlength{\abovecaptionskip}{-0.5cm}
  \caption{\textbf{Overview of DH-VTON.} We demonstrate the training pipeline of our DH-VTON and details of the attention block. \textbf{(Left)} DH-VTON comprises a fixed-parameter PBE\cite{PBE} and a trainable GFC+. 
  Apart from the given noisy image $\mathbf{x}_t$, mask $m$, masked image $\mathbf{x}_0'$, garment image $g$, time steps $t$, GFC+ generates a set of control vectors $c_t$ by incorporating additional control conditions, such as pose $p$ and densepose $d$. 
  Control vectors are integrated into PBE\cite{PBE} to enhance the model's controllability while preserving PBE's generation capabilities. 
  \textbf{(Right)} We introduce a hybrid attention strategy in GFC+ to ensemble different layers of fine-grained characteristics  for multi-scale features preservation.}
  \label{overview}
\vspace{-0.4cm}
\end{figure*}

\textbf{b)} In contrast, another type of warping-free approaches\cite{PBE,MGD,DCI-VTON,LaDI-VTON,tryondiffusion,stableviton,ootdiffusion,IDMVTON,catDM,tpd} usually adopt LDMs\cite{LDM} as image generator because of their strong intrinsic generation capabilities. 
To avoid generating artifacts, they generally bypass garment warping and utilize an feature extractor and several cross-attention blocks to capture and transfer the textures of the given garments. 
For instance, CLIP\cite{CLIP} has emerged as a robust image encoder and is frequently employed as feature extractor in various VTON methods, including MGD\cite{MGD} and PBE\cite{PBE}. 
However, CLIP\cite{CLIP} is pre-trained to align with the holistic features of coarse textual captions~\cite{ma2022cmal,ma2022unitranser}. Therefore, the extracted features are usually also coarse-grained, which may lead to undesirable effects~\cite{ma2023hybridprompt}. 
Recent methods have improved the garment feature extraction abilities by utilizing a tale of two UNet modules, such as TryOnDiffusion\cite{tryondiffusion}, OOTDiffusion~\cite{ootdiffusion}, and IDM-VTON~\cite{IDMVTON}. 
However, these methods still suffer from preserving meticulous details of garments, dampening their applications to real-world scenarios. 

Driven by the above issues, we propose DH-VTON, a deep text-driven virtual try-on model featuring a special hybrid attention learning strategy and deep garment semantic preservation module. 
Specifcally, inspired by the success of PBE\cite{PBE}, we present our DH-VTON pipeline in this work. 
Furthermore, for extracting the deep semantics of the garments, we are the first to introduce InternViT-6B\cite{internvl1.5} into VTON tasks as fine-grained feature learner, which can be trained to align with the large-scale intrinsic knowledge with deep textual semantics to compensate for the deficiency of the commonly adopted CLIP\cite{CLIP} encoder. 
On this basis, to enhance the customized dressing capabilities, we further design GFC+ module and propose to utilize a novel hybrid attention strategy for training, which can adaptively integrate fine-grained characteristics of the garments into the different layers of the VTON model, so as to achieve multi-scale features preservation. 

Experiments on two representative datasets VITON-HD\cite{viton-hd} and DressCode\cite{dresscode} demonstrate the effectiveness of the proposed DH-VTON model, showing that it achieves competitive performance against previous warping-based or warping-free models. 
DH-VTON not only significantly enhances the fine-grained semantic extraction of the given garments but also effectively captures and preserves the texture details. 

\section{Method and Methodology}

To effectively improve the fine-grained semantic extraction abilities and accurately preserve the texture details, we propose a novel deep text-driven virtual try-on (DH-VTON) model, which integrates a special 
hybrid attention strategy and deep garment semantic
preservation module, as depicted in Fig.~\ref{overview}(a). 
DH-VTON mainly contains two parts: a fixed-parameter PBE\cite{PBE} and a trainable GFC+. 
The former aims to ensure high realism of generated images, while the latter aims to further enhance the customized dressing abilities. 
\subsection{ControlNet Architecture}
Given a target person image $\mathbf{x}_0$, DH-VTON gradually adds noise to $\mathbf{x}_0$, receiving a noisy image $\mathbf{x}_t$, with $t$ representing the frequency of noise addition. 
And given a group of conditions including noisy image $\mathbf{x}_t$, mask $m$, masked image $\mathbf{x}_0'$, given garment image $g$, time steps $t$ as well as additional control conditions (\emph{e.g., pose $p$ and densepose $d$}), GFC+ generates a suite of control vectors ${c}_t$. 
Then these vectors are incorporated into the SD Middle Block and the skip-connections of PBE's UNet, consequently guiding the generation process of PBE\cite{PBE}. 
Similar to LDM\cite{LDM}, DH-VTON learns a network $\epsilon_\theta$ to predict the noise added to the noisy image $\mathbf{x}_t$ with: 
\begin{equation}
    \mathcal{L}_{DH-VTON} = \mathbb{E}_{t, \mathbf{x}_\mathrm{0}, \epsilon\sim\mathcal{N}(0, 1)}\left[\lVert\epsilon - \epsilon_{\theta}(\mathbf{x}_t, \mathbf{x}_0', m, g, p, d, t) \rVert_2^2\right],
    \label{eq1}
\end{equation}
where $t\in\{1,...,T\}$ denotes the time step of the forward diffusion process, $\mathbf{x}_\mathrm{0}$ is the target person image and $\mathbf{x}_t$ is $\mathbf{x}_\mathrm{0}$ with the added standard Gaussian noise $\epsilon\sim\mathcal{N}(0, 1)$. 

\subsection{Garment Feature Extraction}
\begin{figure*}
  \centering
  \includegraphics[width=\textwidth]{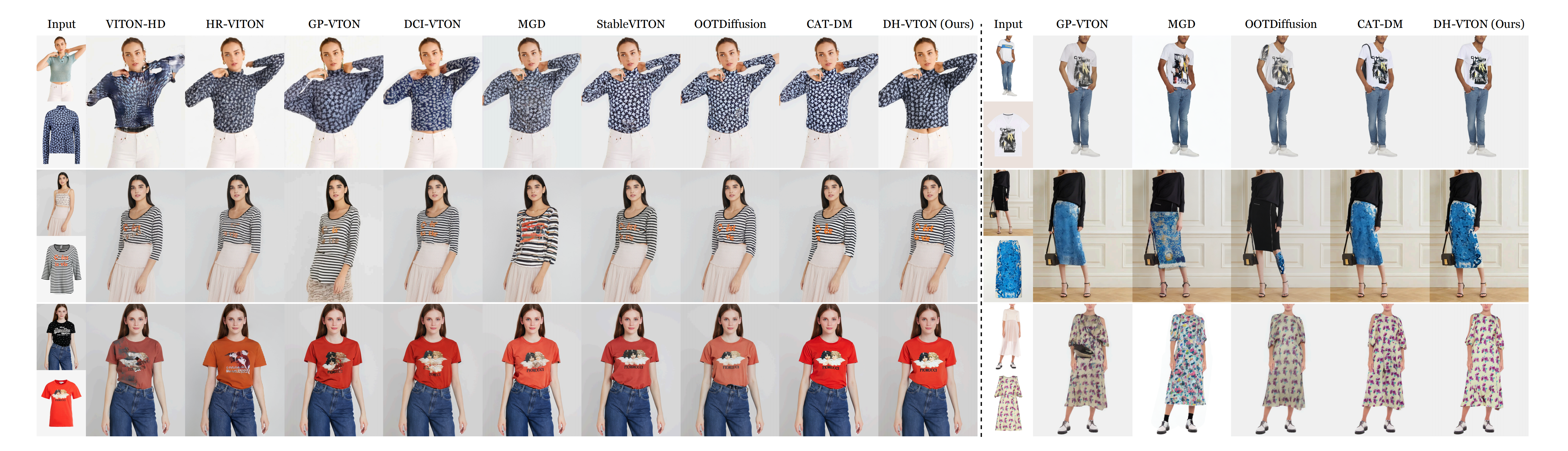}
  \setlength{\abovecaptionskip}{-0.5cm}
  \caption{\textbf{Qualitative results on VITON-HD and DressCode test datasets.} Please zoom in for more details.}
  \label{fig:vitonhd}
\vspace{-0.2cm}
\end{figure*}
\begin{table}[t]
\vspace{-0.25cm}
\newcommand{\tabincell}[2]{\begin{tabular}{@{}#1@{}}#2\end{tabular}}
\setlength{\abovecaptionskip}{0cm}  
\setlength{\belowcaptionskip}{-0.5cm} 
\caption{Ablation study of hybrid attention and different values of $\lambda$ on VITON-HD\cite{viton-hd}. The best and second best results are reported in \textbf{bold} and \underline{underline}, respectively.}
\centering
\renewcommand\arraystretch{1.2}
\scalebox{0.88}{\begin{tabular}{cc|cccc|cc}
\toprule
\multirow{2}{*}{\tabincell{c}{\textbf{Hybrid} \\ \textbf{Attention}}} & \multirow{2}{*}{\textbf{$\lambda$}} & \multicolumn{4}{c|}{Paired} & \multicolumn{2}{c}{Unpaired} \\
\cline{3-8}
& & SSIM\ $\uparrow$  & FID $\downarrow$ & KID $\downarrow$ & LPIPS $\downarrow$  & FID $\downarrow$ & KID $\downarrow$ \\
\hline
\ding{55} & - & 0.795 & 7.63 & 4.21 & 0.1304 & 14.46 & 5.25 \\
\ding{51} & 0.25 & 0.826 & 6.54 & 2.36 & 0.1072 & 12.47 & 3.88 \\
\ding{51} & 0.5 & 0.863 & 5.92 & 0.98 & 0.0861 & 10.92 & 1.62 \\
\ding{51} & 0.75 & \underline{0.871} & \underline{5.58} & \underline{0.71} & \underline{0.0589} & \underline{9.31} & \textbf{1.07} \\
\rowcolor{LightBlue} \ding{51} & 1.0 & \textbf{0.874} & \textbf{5.53} & \textbf{0.67} & \textbf{0.0562} & \textbf{9.02} & \underline{1.08} \\
\ding{51} & 1.25 & 0.859 & 6.12 & 1.96 & 0.0938 & 11.29 & 2.76 \\
\ding{51} & 1.5 & 0.801 & 8.01 & 4.83 & 0.1395 & 15.03 & 6.09 \\
\bottomrule
\end{tabular}}
\label{tab:ablation_lambda}
\vspace{-0.45cm}
\end{table}
\begin{table}[t]
\setlength{\abovecaptionskip}{0cm}  
\setlength{\belowcaptionskip}{-0.5cm} 
\caption{Ablation study of DH-VTON without or with GFC+ on VITON-HD\cite{viton-hd}.}
\centering
\renewcommand\arraystretch{1.2}
\scalebox{0.88}{\begin{tabular}{c|cccc|cc}
\toprule
\multirow{2}{*}{\textbf{GFC+}} & \multicolumn{4}{c|}{Paired} & \multicolumn{2}{c}{Unpaired} \\
\cline{2-7}
& SSIM\ $\uparrow$  & FID $\downarrow$ & KID $\downarrow$ & LPIPS $\downarrow$  & FID $\downarrow$ & KID $\downarrow$ \\
\hline
\ding{55} & 0.763 & 14.32 & 5.44 & 0.2254 & 15.77 & 6.22 \\
\rowcolor{LightBlue} \ding{51} & \textbf{0.874} & \textbf{5.53} & \textbf{0.67} & \textbf{0.0562} & \textbf{9.02} & \textbf{1.08} \\
\bottomrule
\end{tabular}}
\label{tab:ablation_GFC+}
\vspace{-0.4cm}
\end{table}
To make up for the deficiency of the commonly adopted CLIP\cite{CLIP} encoder, we are the first to introduce InternViT-6B\cite{internvl1.5} into VTON tasks as fine-grained feature learner to extract the deep semantics of the garments. 

Specifically, InternViT-6B\cite{internvl1.5} first dynamically matches the image to the optimal aspect ratio from a set of pre-defined aspect ratios. 
Once the appropriate aspect ratio is determined, the image is resized to the corresponding resolution. 
For example, as shown in Fig.~\ref{InternViT}, an $768\times1024$ image is resized to $896\times1344$. 
After that, the resized image is divided into 6 tiles of $448\times448$ pixels and each tile is processed independently. 
In addition to these tiles, a $448\times448$ thumbnail of the entire image is also included to capture the global context for comprehending the overall features. 
\subsection{Hybrid Attention Learning}
In order to adaptively integrate fine-grained characteristics of the garments into the different layers of the VTON model, as shown in Fig.~\ref{overview}(b), we propose to leverage a fresh hybrid attention strategy for training, accordingly achieving multi-scale features preservation. 
Here, assuming $\mathbf{O}_s$ represents the output of self attention and $\mathbf{I}_g$ represents the fine-grained features from InternViT-6B\cite{internvl1.5} at corresponding positions, the output of hybrid attention $\mathbf{O}_h$ can be defined as follows:
\begin{equation}
    \mathbf{O}_{h} = 
    \underbrace{\operatorname{Softmax}\left(\frac{\mathbf{Q} \mathbf{K}^{\top}}{\sqrt{d}}\right) \mathbf{V}}_{\text{Self Attention}} + \lambda\underbrace{\operatorname{Softmax}\left(\frac{\mathbf{Q}\left(\mathbf{K}^{\prime}\right)^{\top}}{\sqrt{d}}\right) \mathbf{V}^{\prime}}_{\text{Cross Attention}} ~,
    \label{eq4}
\end{equation}
where $\lambda\in\mathbb[0, 1.5]$ is a hyper-parameter to control the scale of fine-grained features. 
Note that we share a query matrix $\mathbf{Q}$ for both self attention and cross attention. 
$\mathit{\mathbf{Q}=\mathbf{O}_s\mathbf{W}_q}$, 
$\mathit{\mathbf{K}=\mathbf{O}_s\mathbf{W}_k}$, 
$\mathit{\mathbf{V}=\mathbf{O}_s\mathbf{W}_v}$, 
$\mathit{\mathbf{K}'=\mathbf{I}_g\mathbf{W}'_k}$, 
and $\mathit{\mathbf{V}'=\mathbf{I}_g\mathbf{W}'_v}$. 
Here, $\mathbf{W}_q, \mathbf{W}_k, \mathbf{W}_v, \mathbf{W}'_k$, and $\mathbf{W}'_v$ are the weight matrices of the trainable linear projection layers. 
\begin{table}[t]
\setlength{\abovecaptionskip}{0cm}  
\setlength{\belowcaptionskip}{-0.5cm} 
\caption{Ablation study of DH-VTON with different feature extractors on VITON-HD\cite{viton-hd}.}
\centering
\renewcommand\arraystretch{1.2}
\scalebox{0.88}{\begin{tabular}{l|cccc|cc}
\toprule
\multirow{2}{*}{\textbf{Extractor}} & \multicolumn{4}{c|}{Paired} & \multicolumn{2}{c}{Unpaired} \\
\cline{2-7}
& SSIM\ $\uparrow$  & FID $\downarrow$ & KID $\downarrow$ & LPIPS $\downarrow$  & FID $\downarrow$ & KID $\downarrow$ \\
\hline
CLIP\cite{CLIP} & 0.853 & 7.90 & 1.38 & 0.1111 & 10.21 & 1.77 \\
IP-Adapter\cite{ipadapter} & 0.847 & 8.13 & 2.86 & 0.1127 & 11.23 & 3.90 \\
DINO-V2\cite{dinov2} & \underline{0.862} & \underline{7.11} & \underline{1.12} & \underline{0.0988} & \underline{9.67} & \underline{1.36} \\
\rowcolor{LightBlue} InternViT-6B\cite{internvl1.5} & \textbf{0.874} & \textbf{5.53} & \textbf{0.67} & \textbf{0.0562} & \textbf{9.02} & \textbf{1.08} \\
\bottomrule
\end{tabular}}
\label{tab:ablation_extractor}
\vspace{-0.4cm}
\end{table}
\begin{table*}[t]
\centering
\setlength{\abovecaptionskip}{0cm}  
\setlength{\belowcaptionskip}{-0.4cm} 
\caption{Quantitative results under both paired and unpaired settings on VITON-HD\cite{viton-hd} and DressCode\cite{dresscode} test datasets.}
\renewcommand\arraystretch{1.2}
\scalebox{0.88}{\begin{tabular}{l|cccc|cc|cccc|cc}
\toprule
\multirow{3}{*}{Methods} & \multicolumn{6}{c|}{VITON-HD} & \multicolumn{6}{c}{DressCode} \\
\cline{2-13}
& \multicolumn{4}{c|}{Paired} & \multicolumn{2}{c|}{Unpaired}  & \multicolumn{4}{c|}{Paired} & \multicolumn{2}{c}{Unpaired} \\
\cline{2-13}
& SSIM\ $\uparrow$  & FID $\downarrow$ & KID $\downarrow$ & LPIPS $\downarrow$  & FID $\downarrow$ & KID $\downarrow$ & SSIM\ $\uparrow$  & FID $\downarrow$ & KID $\downarrow$ & LPIPS $\downarrow$  & FID $\downarrow$ & KID $\downarrow$ \\
\hline
VITON-HD\cite{viton-hd} & 0.848 & 12.81 & 5.52 & 0.1216 &  14.64 & 6.10 & - & - & - & - & - & -\\
HR-VITON\cite{HRviton} & 0.860 & 9.92 & 3.06 & 0.1038 &  12.15 & 3.42 & - & - & - & - & - & -\\
DCI-VTON\cite{DCI-VTON} & 0.862 & 9.41 & 4.55 & 0.0606 &  12.53 & 5.25 & - & - & - & - & - & -\\
StableVITON\cite{stableviton} & 0.854 & 6.44 & 0.94 & 0.0905 &  11.05 & 3.91  & - & - & - & - & - & - \\
GP-VTON\cite{GPvton} & 0.871 & 8.73 & 3.94 & \underline{0.0585} & 11.84 & 4.31 & 0.771 & 9.93 & 4.61 & 0.1801 & 12.79 & 6.63 \\
MGD\cite{MGD} & 0.827 & 11.12 & 3.38 & 0.1280 &  13.34 & 3.93 & 0.786 & 8.24 & 3.27 & 0.1078 & 10.19 & 5.84\\
OOTDiffusion\cite{ootdiffusion} & 0.819 & 9.31 & 4.09 & 0.0876 &  12.41 & 4.69  & \textbf{0.885} & 4.61 & \underline{0.96} & \underline{0.0533} & 12.57 & 6.63 \\
CAT-DM\cite{catDM} & \textbf{0.877} & \underline{5.60} & \underline{0.83} & 0.0803 &  \textbf{8.93} & \underline{1.37}  & 0.866 & \underline{4.17} & 1.21 & 0.0674 & \underline{8.22} & \underline{1.98} \\
\rowcolor{LightBlue} DH-VTON (Ours) & \underline{0.874} & \textbf{5.53} & \textbf{0.67} & \textbf{0.0562} &  \underline{9.02} & \textbf{1.08} & \underline{0.881} & \textbf{3.79} & \textbf{0.84} & \textbf{0.0435} & \textbf{6.31} & \textbf{1.36} \\
\bottomrule
\end{tabular}}
\label{tab:effect_comparison}
\vspace{-0.5cm}
\end{table*}
\section{Experiments}
\subsection{Experimental Setup}
\paragraph{Datasets and Metrics} Our experiments are performed on two high-resolution ($768\times1024$) VTON datasets, i.e., VITON-HD\cite{viton-hd} and DressCode\cite{dresscode}. 
And test experiments are conducted under both paired and unpaired settings. 
In the paired and unpaired settings, we employ FID\cite{FID} and KID\cite{KID} for realism and fidelity assessment. 
Furthermore, in the paired setting with available ground truth, we additionally employ LPIPS\cite{LPIPS} and SSIM\cite{SSIM} to evaluate the coherence of VTON images. 
\paragraph{Baselines} For more holistic comparisons, we compare 
DH-VTON with the two categories of baseline models: 1) warping-based models, including VITON-HD\cite{viton-hd}, HR-VITON\cite{HRviton}, GP-VTON\cite{GPvton}, and DCI-VTON\cite{DCI-VTON}; 2) warping-free models, including MGD\cite{MGD}, StableVITON\cite{stableviton}, OOTDiffusion\cite{ootdiffusion}, and CAT-DM\cite{catDM}. 
\paragraph{Implementation Details} During the experiments, we use an end-to-end training process. 
All experiments are conducted on four NVIDIA RTX A6000 GPUs with a batch size of 2. 
We utilize the AdamW optimizer and set the learning rate to $3\times10^{-5}$. 
Moreover, the hyper-parameter $\lambda$ is searched from \{0.25, 0.5, 0.75, 1.0, 1.25, 1.5\}. 
\subsection{Ablation Studies}
\begin{figure}[t]
\vspace{-0.25cm}
  \centerline{\includegraphics{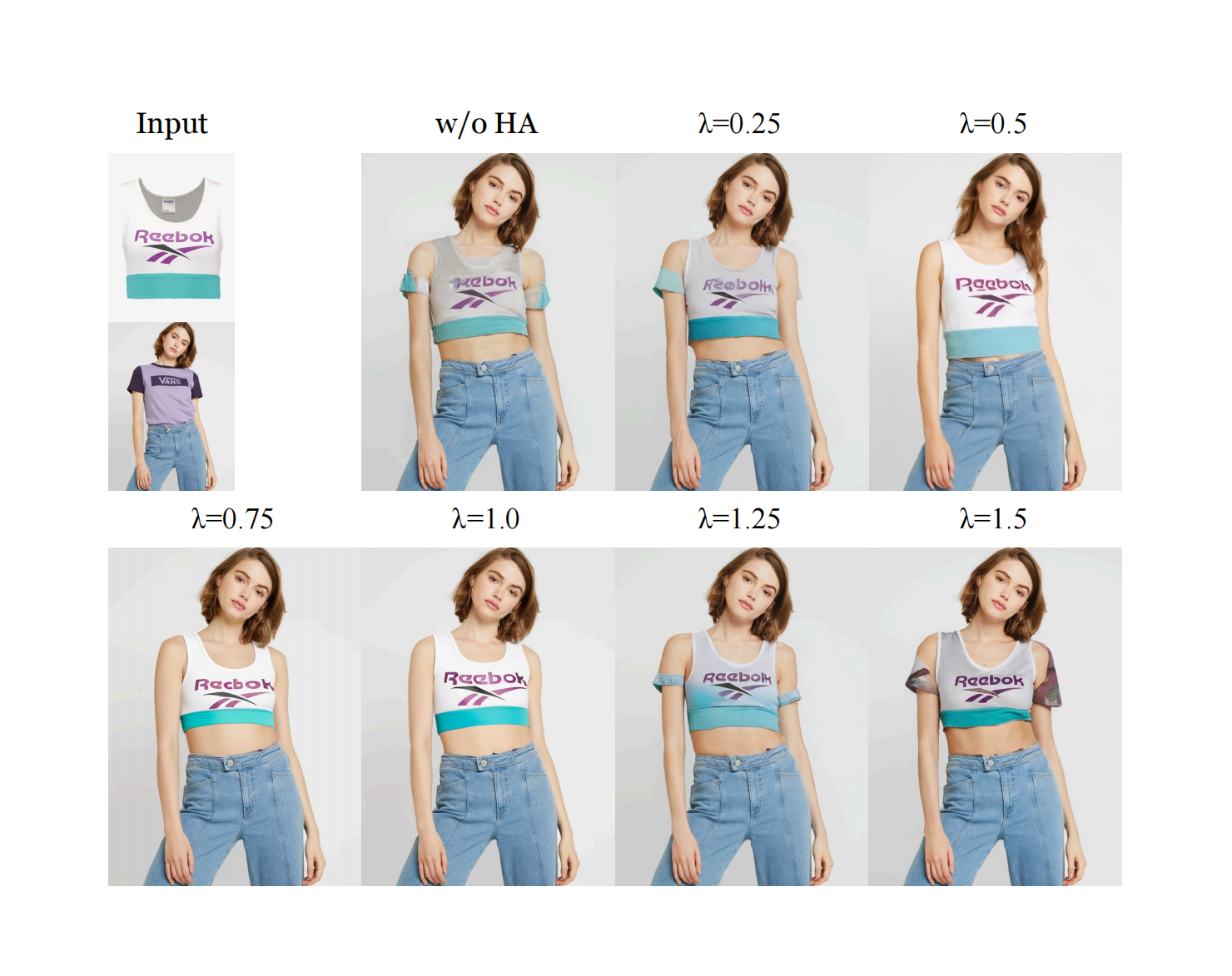}}
  \setlength{\abovecaptionskip}{-0.2cm}
  \setlength{\abovecaptionskip}{-0.5cm}
  \caption{\textbf{Effect of $\lambda$.} We compare the results of DH-VTON trained without/with hybrid attention strategy and using different values of $\lambda$.}
  \label{超参数}
\vspace{-0.45cm}
\end{figure}

\begin{figure}[t]
\centerline{\includegraphics{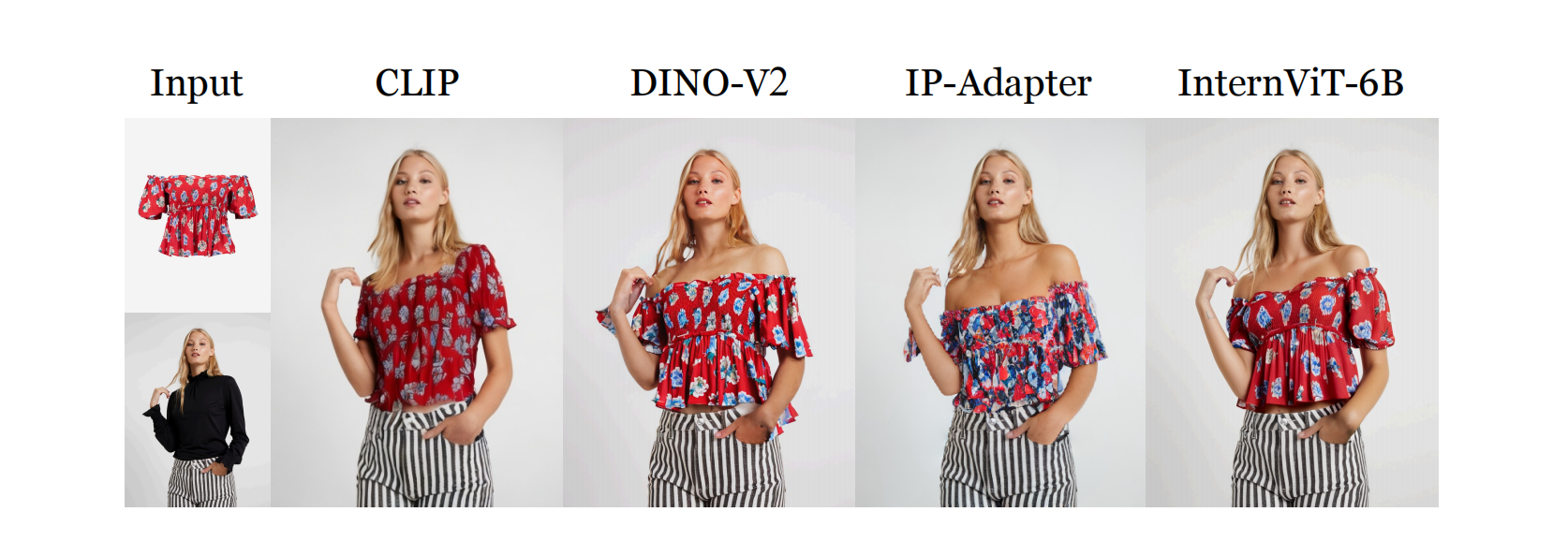}}
\setlength{\abovecaptionskip}{-0.25cm}
\caption{\textbf{Effect of InternViT-6B.} We compare the results of DH-VTON when using different feature extractors.}
\label{特征提取器消融}
\vspace{-0.5cm}
\end{figure}

\paragraph{Hyper-parameter $\lambda$} We investigate the effect of hybrid attention learning as well as the different values of the guidance scale $\lambda$ on VITON-HD\cite{viton-hd}. 
Experimental results are presented in Fig.~\ref{超参数} qualitatively and Tab.~\ref{tab:ablation_lambda} quantitatively. 
We can find that the optimal $\lambda$ value is around 1.0 on VITON-HD\cite{viton-hd}. 
Meanwhile, for more complicated dress images of DressCode\cite{dresscode}, a larger $\lambda$ is needed to match more complex and detailed garment features. 
According to this study, we consistently conduct hybrid attention learning for DH-VTON, and empirically set $\lambda=$ 1.0 for VITON-HD\cite{viton-hd} and $\lambda=$ 1.25 for DressCode\cite{dresscode} in the following experiments. 
\paragraph{Effect of InternViT-6B} We conduct a series of ablation studies to investigate the effect of InternViT-6B\cite{internvl1.5}. Experimental results on how different feature extractors affect the performance of DH-VTON are illustrated in Fig.~\ref{特征提取器消融} qualitatively and Tab.~\ref{tab:ablation_extractor} quantitatively. 
With the integration of InternViT-6B\cite{internvl1.5}, DH-VTON obtains the most realistic and natural VTON results and has shown great progress and improvement across all metrics on VITON-HD\cite{viton-hd}. 

\begin{figure}[t]
\centerline{\includegraphics{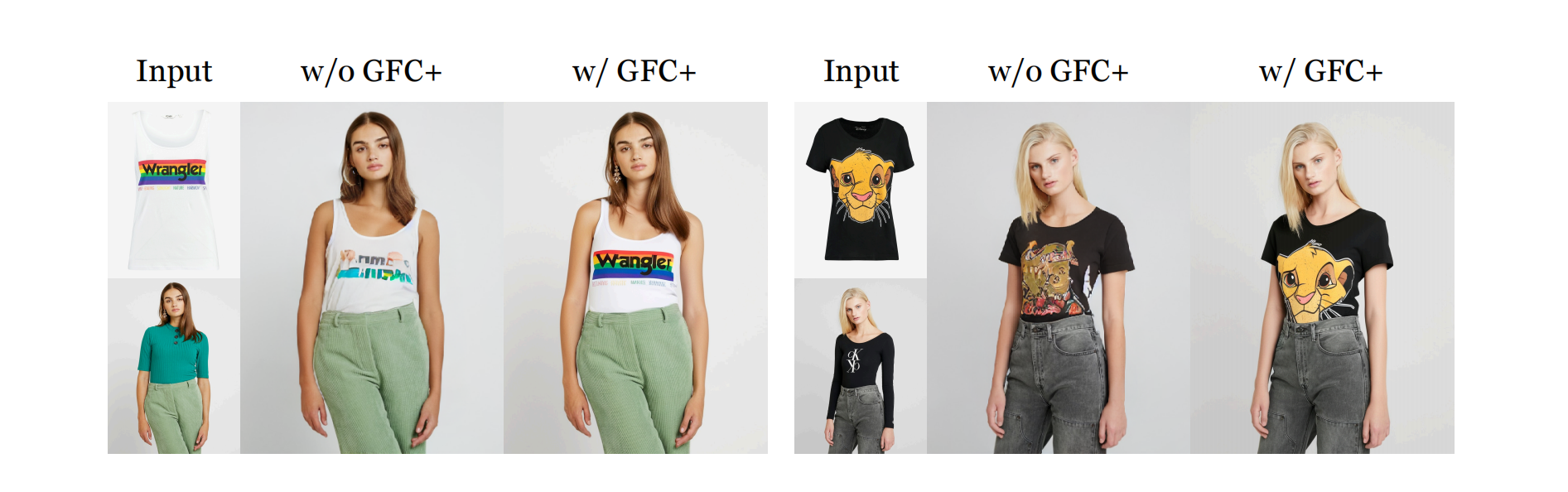}}
\setlength{\abovecaptionskip}{-0.2cm}
\caption{\textbf{Effect of GFC+.} We compare the results of DH-VTON trained without/with GFC+.}
\label{GFC+消融}
\vspace{-0.6cm}
\end{figure}

\paragraph{Effect of GFC+} We also investigate the effect of GFC+ on VITON-HD\cite{viton-hd}. 
Experimental results are shown in Fig.~\ref{GFC+消融} qualitatively and Tab.~\ref{tab:ablation_GFC+} quantitatively. 
GFC+ visually enhances the customized dressing abilities of preserving the textures and patterns of the given garments (\emph{e.g., the texts and graphics of t-shirts}) and quantitatively improves all evaluation metrics, which consistently shows the superior of our model. 
\subsection{Qualitative Results}
Some test results of DH-VTON compared to other VTON methods on VITON-HD\cite{viton-hd} and DressCode\cite{dresscode} datasets are visually shown in Fig.~\ref{fig:vitonhd}. 
Compared with other methods, we can observe that our DH-VTON significantly achieves the best try-on effect for various kinds of garments. 
Moreover, our DH-VTON not only generates realistic images but also preserves most of the fine-grained garment details. 
\subsection{Quantitative Results}
The quantitative comparisons between DH-VTON and other methods are minutely reported in Tab.~\ref{tab:effect_comparison}. 
DH-VTON outperforms other methods on the majority of metrics, particularly in KID\cite{KID} and LPIPS\cite{LPIPS}, demonstrating its effectiveness in image generation quality on both paired and unpaired tasks. 
\section{Conclusion}
In this paper, we present DH-VTON, a deep text-driven virtual try-on model. 
We are the first to introduce InternViT-6B\cite{internvl1.5} into VTON tasks as fine-grained feature learner, which significantly improves the deep semantic extraction abilities. 
Besides, we make full use of inherent power within PBE\cite{PBE} and design an additional GFC+ module to enhance the customized dressing abilities. Based on this, we further propose a fresh hybrid attention strategy to ensemble different layers of fine-grained characteristics for multi-scale features preservation. 
Experiments on the two representative datasets demonstrate the effectiveness and superior of our approach.


\bibliographystyle{IEEEtran}
\bibliography{IEEEabrv,ref}

\end{document}